\documentclass{article}

\usepackage{PRIMEarxiv}
\usepackage{algorithm}
\usepackage[noend]{algpseudocode}
\usepackage{multirow}
\usepackage[utf8]{inputenc} 
\usepackage[T1]{fontenc}    
\usepackage{hyperref}       
\usepackage{url}            
\usepackage{booktabs}       
\usepackage{amsfonts}       
\usepackage{nicefrac}       
\usepackage{microtype}      
\usepackage{lipsum}
\usepackage{fancyhdr}   
\usepackage{graphicx} 
\usepackage{float}
\usepackage{color}
\graphicspath{{media/}}    

\title{Going Forward-Forward in Distributed Deep Learning}

\author{
  \textbf{Ege Aktemur*, Ege Zorlutuna, Kaan Bilgili, Tacettin Emre Bök, Berrin Yanikoglu, Süha Orhun Mutluergil}  \\
  Faculty of Engineering and Natural Sciences, Sabanci University, Istanbul Türkiye 34956 \\
  \texttt{\{egeaktemur, egezorlutuna, kaanbilgili, tacettinemre, berrin, suha.mutluergil\}@sabanciuniv.edu}\\
}

\begin{document}
\maketitle

\begin{abstract}
We introduce a new approach in distributed deep learning, utilizing Geoffrey Hinton's Forward-Forward (FF) algorithm to speed up the training of neural networks in distributed computing environments. Unlike traditional methods that rely on forward and backward passes, the FF algorithm employs a dual forward pass strategy, significantly diverging from the conventional backpropagation process. This novel method aligns more closely with the human brain's processing mechanisms, potentially offering a more efficient and biologically plausible approach to neural network training. 
Our research explores different implementations of the FF algorithm in distributed settings, to explore its capacity for parallelization.
While the original FF algorithm focused on its ability to match the performance of the backpropagation algorithm, the parallelism aims to reduce training times and resource consumption, thereby addressing the long training times associated with the training of deep neural networks. 
Our evaluation shows a 3.75 times speed up on MNIST dataset without compromising accuracy when training a four-layer network with four compute nodes.
The integration of the FF algorithm into distributed deep learning represents a significant step forward in the field, potentially revolutionizing the way neural networks are trained in distributed environments.
\end{abstract}

\section{Introduction}
\begin{figure}[thb]
  \centering
  \includegraphics[width=0.8\textwidth]{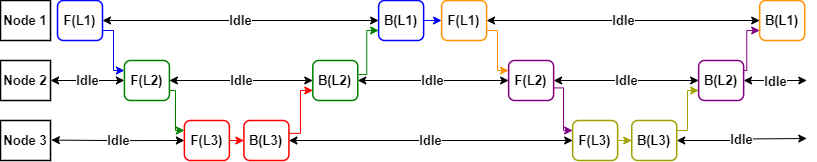}
  \caption{Challenges in parallelizing traditional backpropagation involve managing the forward (\textit{F}) and backward (\textit{B}) passes, both applied to the layer specified by their first parameter.}
  \label{fig:distributed_bp}
\end{figure}

Training deep neural networks, often consisting of  over hundreds of layers, is a very time consuming process that can take several weeks, using the well-known backpropagation \cite{Rumelhart1986Backprop} as the learning algorithm.
On the other hand, parallelization of this algorithm presents significant challenges due to the sequential nature of the backpropagation learning algorithm, as illustrated in Figure \ref{fig:distributed_bp}. 

In backpropagation, the gradient of each layer depends on the gradient calculated at the next layer. In a distributed setting, this sequential dependency necessitates extensive waiting times between computing nodes, as each node must wait for the gradient information to backpropagate from its successor before it can proceed with its calculations. 
Moreover, the need for constant communication between nodes to transfer gradient and weight information can lead to significant communication overhead. This is particularly problematic in large-scale neural networks where the volume of data to be transferred can be substantial. 

The realm of distributed deep learning has witnessed significant advancements in recent years, driven by the ever-increasing complexity and size of the neural networks. Distributed training frameworks like GPipe \cite{huang2019gpipe}, PipeDream \cite{pipedream} and Flower \cite{flower2022federated} have emerged as pivotal solutions, enabling the training of massive models by optimizing for scale, speed, cost, and usability. These systems utilize sophisticated methodologies such as data, pipeline and model parallelism to manage and process large-scale neural network training efficiently across multiple computing nodes.

\begin{figure}[thb]
  \centering
  \includegraphics[width=0.55\linewidth]{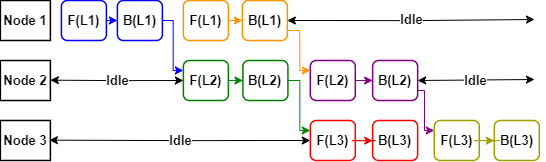}
  \caption{Parallelization using the FF algorithm.}
  \label{fig:distributed_ff}
\end{figure}

Aside from the above research geared towards distributed implementations of backpropagation, the Forward-Forward (FF) algorithm proposed by Hinton \cite{hinton2022forward} proposes a novel approach for training neural networks. Unlike traditional deep learning algorithms that rely on global forward and backward passes, the Forward-Forward algorithm only uses local (layer-wise) computations. 
While FF algorithm is designed to address the limitations associated with backpropagation without the focus of distributing, layer-wise training feature of FF results in a less dependent architecture in distributed setting that reduces idle time, communication and synchronization as demonstrated in Figure \ref{fig:distributed_ff}. 

This paper explores the integration of the FF algorithm into distributed deep learning systems, particularly focusing on its potential to improve the efficiency and effectiveness of training large-scale neural networks. 
We demonstrate the effectiveness of the Pipeline Forward-Forward (PFF) algorithm, which reduces the training duration by almost 5-fold for a 5-layer network while preserving the accuracy of the original algorithm.

\section{Literature Review}
Distributed and parallel neural networks have attracted widespread interest and investigation from researchers, driven by the promise of improved computational efficiency and scalability in handling large-scale data and complex models. We used several comprehensive surveys from 2019 \cite{ben2019demystifying} and 2022 \cite{mayer2020scalable}. These surveys explain the parallelism methods including data, model and pipeline that we employ in our methods.  

PipeDream \cite{pipedream} is a deep neural network training system using backpropagation that efficiently combines data, pipeline, and model parallelism by dividing the layers of the model into multiple stages that contains multiple consecutive layers. Unlike our method, PipeDream inherits the limitations of the classic back propagation algorithm.
Similar to PipeDream, GPipe \cite{huang2019gpipe} is a distributed deep learning library that updates the weights synchronously during backward pass, which differentiates it from PipeDream. However, it also suffers from additional dependencies introduced by the classic back propagation algorithm.

Besides the above methods, there are several proposed methods for efficient deep learning training that attempts to overcome the limitations of back propagation by proposing alternatives to back propagation. One of them is called \textit{local parallelism} \cite{laskin2021parallel}. This method divides the layers into number of blocks, and use this block's local loss to adjust the parameters of the layers that are within that block. Even though this method achieves a significant speed up compared to back propagation, it is effective to an extent because the local loss of the blocks are not proportional to global loss. 

Forward-Forward algorithm \cite{hinton2022forward} is a learning method for multi-layer networks, proposed by Hinton. This  algorithm introduces a novel way of training, which eliminates the back-propagation and backward pass. Different from our method, Forward-Forward Algorithm is sequential, and is designed to work on a single processing unit. 
This algorithm forms the backbone of our method along with the split method, which aims to reduce communication between workers and increase parallelism.
The Forward-Forward algorithm has impacted the field and led to several subsequent works, including neural systems combining Forward-Forward and Predictive Coding \cite{ororbia2023predictive} and training CNNs with closer accuracy (99.20 accuracy better then the Original FF \cite{hinton2022forward}) to backpropagation using FF. Furthermore, CaFo \cite{zhao2023cascaded} which is an algorithm that uses FF Algorithm to create multiple cascaded neural blocks which consists of several different layers. In their paper, they also state that the training of different blocks is independent from each other and suitable for parallelization. 

Distributed Forward-Forward (DFF) \cite{deng2023dff} is an example of the use of FF algorithm in distributed training system. DFF aims to create a decentralized and distributed training system to allow multiple low-performance devices to train a large model. DFF creates a cluster with one master node and multiple server nodes. Each server is assigned layer(s) by the master node. These server nodes train their assigned layers using FF Algorithm and send their output to other layers. 

While PipeDream and GPipe aim to parallelize backpropagation, DFF \cite{deng2023dff} is the closest to our work as it also tries to integrate the FF algorithm into a distributed setting. However, in terms of accuracy, our implementation is far superior than DFF because, our implementation updates the weights more frequently, uses minibatches, generates the negative samples adaptively, and uses an additional classifier which all contribute to the performance of the network.

\section{Forward-Forward Algorithm}
The FF algorithm, proposed by Geoffrey Hinton \cite{hinton2022forward}, presents a novel approach to training neural networks that doesn't involve backpropagation, which is the standard learning algorithm used in training neural networks. 

Inspired by the biological processes of the brain, the FF algorithm trains layers individually and sequentially, utilizing two forward passes to update the layer weights. Those passes are called respectively, the positive pass and the negative pass. The positive pass adjusts the weights to increase the ''goodness'' of the positive or real data, while the negative pass does the opposite for negative data, as defined below. 

\paragraph{Negative Data.}
In \cite{hinton2022forward}, positive and negative samples are generated based on the existence of a black border in the MNIST  dataset which is used in evaluations. A positive sample is generated by adding the correct label on the top-left corner of the input image (a 10-pixel area in the border denotes the label where a 1-of-C encoding is used),  whereas a negative sample is one where a wrong label is selected. 
Other alternatives for creating negative data, including unsupervised methods, are also possible and explored in \cite{hinton2022forward}. 

\paragraph{Goodness Function.}
The goodness of a layer explored in \cite{hinton2022forward} is the sum of the squares of the activities of the rectified linear neurons in that layer and the aim of the learning is to make it above some threshold value for positive (real) data and below the threshold for negative data. This can be captured by applying the logistic function to the sum of squared activities $y_j^2$ minus a threshold $\theta$:
\begin{equation}
p(real)  = \sigma(\Sigma_j y_j^2 - \theta)   
\end{equation}

\paragraph{Prediction.}
Two methods have been proposed for predicting the label of an input image. 
In the \textit{Goodness} prediction, the input is run with 10 different labels where the target vector of a single 1 is placed in the correct class is indicated and the activations from all but the first hidden layer of the network are accumulated. Then, the label resulting in the maximum goodness is  selected as the predicted class. 

In the \textit{Softmax} prediction, the unlabelled input image is labelled with a neutral label that consists of 0.1 values for each label and the  activity vectors collected from all but the first hidden layer are fed into a Softmax layer which is trained to predict the class. 
The softmax layer is trained using backpropagation either at the end of the training or during the training.
To remark since the softmax layer does not propagate to the FF layers, it can also be trained separately like a FF layer.
This prediction approach is faster for inference as there is only a single pass; on the other hand, it is sub-optimal to the Goodness prediction which better corresponds to how the network was trained.  

\paragraph{Sequential Training of the Forward-Forward Algorithm.}
The layers of the network are trained for several epochs where an epoch means passing the whole training data from the network once. Since FF does not involve backpropagation, layers can be trained independently, each layer can be trained for more than one epoch waiting the upper layers to finish the previous epochs. For instance, assuming that the network will be trained for 60 epochs, we can train the first layer for 30 epochs, and use this half-trained layer-1 for training the second layer for its 30 epochs and so on. 

Moreover, training a layer for total number of epochs and providing its output to the next layer is not good for accuracy. Training each layer for small number of epochs allow fine grained adjustments on this layer and yields better results. So, we divide training process of each layer into splits, basically a continuous number of undivided training number of epochs for this layer. 

\begin{figure}[thb]
  \centering
  \includegraphics[width=0.4\textwidth]{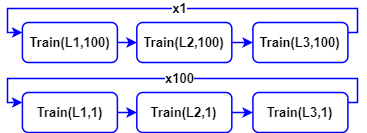}
  \caption{Comparison of the Forward-Forward algorithm with split 1 and split 100 (S = 1, 100) across three layers.}
  \label{fig:Traditional_ff}
\end{figure}

\newpage
\section{PipelineFF: A Pipeline Forward-Forward Algorithm}
In this section we propose  three  Pipeline Forward Forward (PFF) algorithms: Single-Layer PFF (\ref{single-layerPFF}), All-Layers PFF (\ref{all-layersPFF}), Federated PFF (\ref{federatedPFF}) and Performance-Optimized PFF (\ref{within}).  

The Single-Layer PFF employs an architecture where each node is dedicated to training only one layer, while 
All-Layers and Federated PFF algorithms distribute the training load more evenly, by having every node train all network layers in turn. 
All models enjoy model and pipeline parallelism, while the federated PFF also takes advantage of data parallelism.

In all of the proposed pipelined FF models, we split the total number of epochs $E$ into the total number of the splits or chapters \(S\) to facilitate distributed training. 
Each chapter thus takes $C=E/S$ training epochs.   
Consequently, each client will process \(S\) chapters to complete the training for \(E\) epochs.
The variable \(N\) denotes the number of nodes in the distributed system.
The input dataset is abstracted as \(x\), which is a concatenated form of the positive and negative datasets, simplifying the explanation. 
\(L_i(x)\) is used to denote the output of the \(i\)-th layer of the neural network after a forward pass with the input data \(x\). 
Finally, \(Train(L_i, n)\) is an abstracted function that trains layer \(L_i\) for \(n\) epochs.
\subsection{Single-Layer PFF Algorithm}
\label{single-layerPFF}
\begin{figure}[H]
  \centering
  \includegraphics[width=0.8\textwidth]{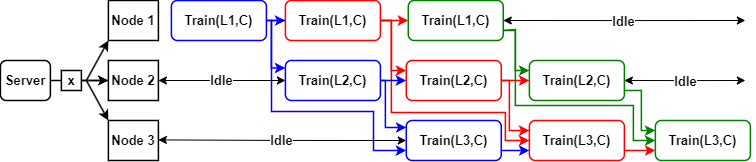}
  \caption{Single-layer Pipeline Forward-Forward Algorithm example with three layers and three splits.}
  \label{fig:Single-Layer_Distributed_ff}
\end{figure}

Single-Layer PFF employs an architecture where each node is dedicated to training only one layer, as shown in Fig. \ref{fig:Single-Layer_Distributed_ff}.
This procedure ensures that while one node trains a layer, the previous node can simultaneously train the preceding layer. 
Already with this simple architecture, we achieve parallelism among the nodes, unlike the sequential FF algorithm.

The distributed training example provided in Figure \ref{fig:Single-Layer_Distributed_ff} illustrates a scenario with three compute nodes and a three-layer neural network. 
The number of nodes, $N$, is selected to be equal to the number of layers $L$ and each node trains the corresponding layer of the network. 
Below, we detail the algorithm with a pseudo code and the training process.

\textbf{Node Initialization.} Each node in the distributed computing system, designated as Node \(i\) where \(i\) ranges from 1 to \(N\), is initialized with the necessary data (training data and hyperparameters) to begin training its corresponding layer \(L_i\).

\textbf{Initial Training:} \(Node_{1}\) begins by training the randomly initialized first layer \(L_1\) using \(x\) for \(C\) epochs.

\textbf{Rest of the Layers:} Concurrently, as \(Node_{i-1}\) completes its \(C\) epochs \(L_{i-1}\), \(Node_{i}\) commences its training of the \(L_{i}\), using the output of \(L_{i-1}(x)\). 

\textbf{Iteration and Convergence:} The process iterates for \(S\) chapters, with each node refining its layer based on the progressively improved outputs from the preceding layer. Convergence is achieved when All-Layers have been trained for \(E\) epochs across all splits.

\begin{algorithm}
\caption{Single-Layer Training Procedure}
\begin{algorithmic}[1]
\For{$chapter$ = $1$ to $split$}
    \State $x_{pos}, x_{neg} \gets X_{POS}, X_{NEG}$
    \For{$layerIndex$ = $1$ to  $currentClientIndex$}
        \State $layer \gets getLayer(layerIndex, chapter)$
        \State $x_{pos}, x_{neg} \gets layer(x_{pos}, x_{neg})$
    \EndFor
    \For{$miniEpoch$ = $1$ to $epochs / split$}
        \State $learningRateCooldown(chapter, miniEpoch)$
        \For{each $batch$ in $x_{pos}, x_{neg}$}
            \State $trainLayer(currentClientIndex, batch)$
        \EndFor
    \EndFor
    \State $PublishLayer(chapter, layerIndex)$
    \State $UpdateXNEG(publish = False)$
\EndFor
\end{algorithmic}
\end{algorithm}

\newpage
\subsection{All-Layers PFF Algorithm}
\label{all-layersPFF}

\begin{figure}[H]
  \centering
  \includegraphics[width=0.9\textwidth]{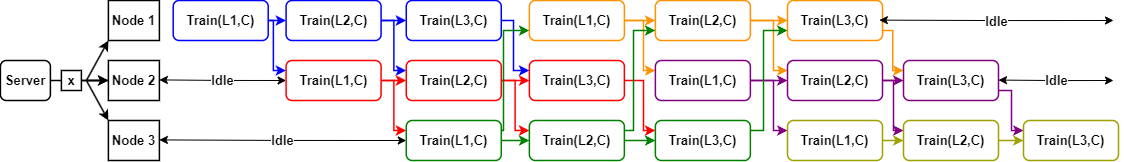}
  \caption{All-Layers Pipeline Forward-Forward Algorithm example with three layers and six splits.}
  \label{fig:AllLayersDistributed_ff}
\end{figure}

The All-Layers PFF model builds upon the Single-Layer approach by allowing each compute node to train the entire network, affording a more balanced load among the nodes. 

Figure \ref{fig:AllLayersDistributed_ff} illustrates the approach in the context of a system with three compute nodes and a three-layer neural network, but with six splits for clarity.

\textbf{Node Initialization.} Each node is initialized with the dataset and the required hyper-parameters. 

\textbf{Initial Training:} \(Node_{1}\) trains the entire network with randomly initialized layers for \(C\) epochs.

\textbf{Rest of the training:} Concurrently, each \(Node_i\) trains layer \(L_{k-1}\) immediately after \(Node_{i-1}\) has completed its \(C\) epochs of training for layer \(L_{k-2}\). Each \(Node_{i}\) sends its last trained layer to \(Node_{i+1(mod(N))}\) and gets the layer to be trained from \(Node_{i-1(mod(N))}\). 

\textbf{Iteration and Convergence:} The process repeats for \(S/N\) but not \(S\) since in this approach every node trains each layer. The algorithm converges when the entire network has been trained on total of \(E\) epochs.

\begin{algorithm}
\caption{All-Layers Training Procedure}
\begin{algorithmic}[1]
\For{$chapter$ = $1$ to $split$}
    \State $x_{pos}, x_{neg} \gets X_{POS}, X_{NEG}$
    \For{$layerIndex$ = $1$ to $numLayers$} 
        \If{not ($firstClient$ and $chapter = 0$)}
            \State $getLayer(layerIndex, chapter)$
        \EndIf
        \For{$miniEpoch$ = $1$ to $epochs / split$}
            \State $learningRateCooldown(chapter, miniEpoch)$
            \For{each $batch$ in $x_{pos}, x_{neg}$}
                \State $trainLayer(layerIndex, batch)$
            \EndFor
        \EndFor
        \State $PublishLayer(chapter, layerIndex)$
        \State $x_{pos}, x_{neg} \gets layer(x_{pos}, x_{neg})$
    \EndFor
    \State $UpdateXNEG(publish = False)$
\EndFor
\end{algorithmic}
\end{algorithm}

\subsection{Federated PFF Algorithm}
\label{federatedPFF}
In contrast to the All-Layers PFF Algorithm, the Federated PFF Algorithm employs the same training procedure but differentiates in the data it utilizes for training. Each node trains on its own local dataset rather than a shared one. This preserves data privacy and autonomy as there is no need to share raw data with other nodes or a centralized server. 
While the data remains localized, nodes exchange model updates. 
This allows nodes to benefit from learning carried out by their counterparts, which iteratively improves the global model  that benefits from diverse data without central aggregation. 
This federated structure is not only crucial for scenarios requiring data privacy but also enables distributed training over networks where central data aggregation is impractical or undesired.


\begin{figure}[thb]
  \centering
  \includegraphics[width=0.8\linewidth]{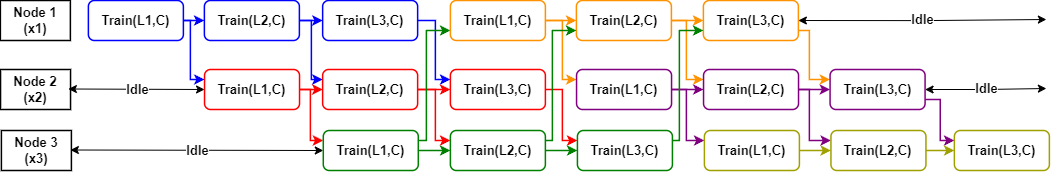}
  \caption{Federated Forward-Forward Algorithm example with 3 layers and 6 split.}
  \label{fig:Federated_ff}
\end{figure}

\newpage
\subsection{PFF with A New Goodness Function}
\label{within}

The Forward-Forward algorithm \cite{hinton2022forward} states that different goodness functions may be used while training each layer individually. In this section, we propose a new PFF algorithm that uses a goodness function based on classification accuracy. This model is called Performance-Optimized PFF.

\begin{figure}[thb]
  \centering
  \includegraphics[width=0.3\linewidth]{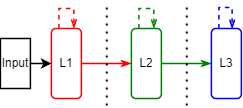}
  \caption{Forward and Backward passes of original FF}
  \label{fig:FF_train}
\end{figure}

Specifically, we add a softmax layer to the network layer that is trained with the PFF algorithm and update that layer's weights using backpropagation for only the two layers (the added softmax and the actual layer), as illustrated in Figure \ref{fig:FF_train}. 
In other words, we train each layer to maximize the classification accuracy with the help of an added softmax layer; hence, this approach is only suitable for supervised learning problems.
Furthermore there is no negative data, but the rest of the training is done exactly in accordance with the FF algorithm.

\begin{figure}[thb]
  \centering
  \includegraphics[width=0.3\linewidth]{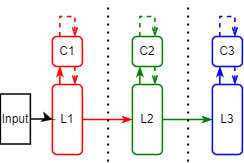}
  \caption{Forward and Backward passes of the proposed FF algorithm, with a separate classifier block for each layer.}
  \label{fig:softmax_train}
\end{figure}

\section{Experimental Evaluation}

We obtain and compare several models by modifying the distributed model (Single-Layer or All-Layers), negative data selection (Adaptive, Fixed or Random), classification strategies (Goodness and Softmax) and goodness function.

There are two PFF variants: Single-Layer and All-Layers models, as explained in Section ~\ref{single-layerPFF} and Section ~\ref{all-layersPFF} respectively. These variants are compared to the sequential implementation (node count $N=1$), which is equivalent to the original FF algorithm and can be used for comparing the models using the same code base.

For selecting negative samples, we implemented AdaptiveNEG, the method proposed in \cite{hinton2022forward}, which selects the most predicted incorrect label as the negative label. 
The second approach is FixedNEG, which involves choosing random incorrect labels for each instance {in the beginning} of the training. The third approach, called RandomNEG, selects the random incorrect labels at the end of each chapter. The last two are proposed to speed up the training. 

For the classification approach, we implemented both approaches proposed in  \cite{hinton2022forward}. In the Goodness based approach, a sample is classified into the class that results in the highest goodness score (considering all but the first layer). In the Softmax approach, a single layer classification head is added to the network with the softmax activation. 

Finally, we evaluated a new goodness function for supervised learning scenarios.

\subsection{Implementation Details}
Our chosen architecture is the same feed-forward neural network as in \cite{hinton2022forward}. 
Its configuration is [784, 2000, 2000, 2000, 2000], where the numbers indicate the number of nodes in each layer. 
Thus the input, which is 784-dimensional, is followed by four hidden layers with 2000 nodes each. 
Each of these layers employs the ReLU activation function.

The network is trained and tested using the MNIST dataset, using 60,000 training instances with mini-batches of size 64 and testing on a separate test set comprising 10,000 instances for 100 epochs and 100 splits. 
The Adam optimizer is used for both the FF Layers and the Softmax layer (trained using Backpropagation) is set initially at 0.01 and 0.0001 respectively and cooldowns after the 50th epoch.
The threshold coefficient $\theta$ in Eq. 1 is set to 0.01, as in \cite{hinton2022forward}. 

\subsection{Comparison of FF, DFF and PFF Models}
\label{main}

We first compare the accuracy and speed of the PFF variants using the Goodness approach, with the results of the original Forward-Forward paper's Matlab implementation \cite{hinton2022forward} and another distributed implementation \cite{deng2023dff} of the Forward-Forward algorithm. The results are given in  Table \ref{tab:main}.

The AdaptiveNEG approach, which selects negative data based on the network's performance at each chapter, outperforms both RandomNEG and FixedNEG models in terms of accuracy. Remarkably, despite shorter training times (7,178 vs 11,190 seconds), RandomNEG performs close to AdaptiveNEG in terms of accuracy (98.33 vs 98.52 in Sequential), making these models more accurate than FixedNEG in every implementation.

The AdaptiveNEG Goodness model with Sequential implementation is basically the same as \cite{hinton2022forward}  
and it indeed achieves almost the same accuracy as Hinton`s Matlab implementation \cite{origpythonimp}.
It enables us to compare the training time of the model with our Python implementation. 

Furthermore, AdaptiveNEG with All-Layers implementation matches the top accuracy of its sequential version (98.51\% vs. 98.52\%) but in about a quarter of the time (2,980 vs. 11,190 seconds), using four compute nodes. This finding showcases the PFF Algorithm's efficiency in distributing the training, achieving 94\% utilization (3.75/4), significantly speeding up the process. 
The All-Layers method speeds up training for AdaptiveNEG by efficiently distributing tasks between nodes, allowing each to compute its own negative samples after every chapter, unlike the Single-Layer approach where the last node generates and publishes the generated labels.

\begin{table}[thb]
\centering
\medskip
\caption{Original, DFF and PFF comparison. Baseline is highlighted as bold.}
\begin{tabular}{|l|l|r|r|}
\hline
\textbf{Model}              & \textbf{Implementation}   & \textbf{Training Time (s)}    & \textbf{Test Accuracy (\%)}   \\ \hline
{Hinton`s Matlab Code \cite{origpythonimp}}     &       &                               & {98.53}                       \\ \hline
{DFF (1000 epochs)\cite{deng2023dff}}           &       &                               & {93.15}                       \\ \hline
                            & {Sequential}              & {11,190.72}                   & {98.52}                       \\ 
{AdaptiveNEG-Goodness}      & {Single-Layer}            & 5,254.87                      & 98.43                         \\ 
                            & {All-Layers}              & \textbf{2,980.76}             & \textbf{98.51}                \\ \hline
                            & {Sequential}              & 7,178.71                      & 98.33                         \\ 
RandomNEG-Goodness          & {Single-Layer}            & 1,974.10                      & 98.26                         \\ 
                            & {All-Layers}              & 2,008.25                      & 98.17                         \\ \hline
                            & {Sequential}              & 7,143.28                      & 97.95                         \\ 
FixedNEG-Goodness           & {Single-Layer}            & 1,920.80                      & 97.94                         \\ 
                            & {All-Layers}              & 1,978.21                      & 97.89                         \\ \hline
\end{tabular}
\label{tab:main}
\end{table}

\subsection{Classifier Mode Comparison for AdaptiveNEG}
\label{Adaptive}
In the first evaluation, we compared PFF variants that used the Goodness approach for classification and established the AdaptiveNEG Goodness model as the proposed baseline. 
The second experiment compares the performance of this model with the one using the Softmax approach.
As mentioned before, the goodness method is the first method provided in \cite{hinton2022forward} and uses only the FF layers. The Softmax method on the other hand is presented as a faster alternative, predicting the output using a Softmax layer that gets the activations of the hidden FF layers (except the first). 
The corresponding  results can be found in Table \ref{tab:adaptive}. 

The results show that the AdaptiveNEG Softmax model trains faster than the AdaptiveNEG-Goodness model, for all different implementations, with time savings ranging from 25-53\% 
(e.g. 1886.42 versus 2980.76 secs).
This accelerated training is due to the single-step prediction in the softmax approach, rather than predicting across 10 different classes. 
However, the accuracies all slightly decrease compared to comparable Goodness models.

\begin{table}[thb]
\centering
\medskip
\caption{Classification mode comparison for AdaptiveNEG. The first three row results indicated in italic denote the baseline model from Table 1. Bold results indicate suggested models for each classification mode.}
\begin{tabular}{|l|l|r|r|r|}
\hline
\textbf{Model}          & \textbf{Implementation}   & \textbf{Training Time (s)}    & \textbf{Test Accuracy (\%)}   \\ \hline
                        & Sequential                & 11,190.72                     & 98.52                         \\
\textit{AdaptiveNEG-Goodness}  & Single-Layer              & 5,254.87                      & 98.43                         \\       
                        & {All-Layers}              & \textbf{2,980.76}             & \textbf{98.51}                \\ \hline
                        & Sequential                & 8,365.96                      & 98.38                         \\
{AdaptiveNEG-Softmax}   & Single-Layer              & 2,471.27                      & 98.31                         \\       
                        & All-Layers                & \textbf{1,886.42}                      & \textbf{98.30}                         \\ \hline
\end{tabular}
\label{tab:adaptive}
\end{table}

\subsection{Classifier Mode Comparison for RandomNEG}
\label{other}

Building upon previous findings in the section \ref{main}, we observed that RandomNEG closely rivals AdaptiveNEG in accuracy, while significantly reducing training time. This observation leads us to further investigate the impact of employing softmax within the RandomNEG framework. This section thus compares the effects of softmax and Goodness prediction on both FixedNEG and RandomNEG strategies. The results of this comparison are detailed in Table \ref{tab:smax-random}, offering insights into how these elements influence the effectiveness and efficiency of the methods under review.

In terms of accuracy, the Softmax approach outperforms the Goodness approach for prediction. For instance, while Sequential RandomNEG Goodness model performed 98.33 the softmax version performed 98.48. This accuracy level of the softmax version is very close to the performance of our baseline model (98.51 Accuracy using AdaptiveNEG Goodness). While the accuracy levels are very close we observe that there is a significant speed up using RandomNEG (8,104 versus 11,190).

In terms of training speed on the other hand, training a model with the additional Softmax layer takes longer in the Sequential model as expected. However, Single-Layer implementation is not effected from this additional complexity. This is due to the fact that Softmax Layer is easier/faster to train than the FF Layers and thanks to the pipeline architecture this only adds a small delay. 

Surprisingly, Softmax models give faster results compared to same Goodness models for All-Layers implementation. This again stems from the fact that training Softmax is faster than training FF Layers and distributing this job between 5 nodes speeds ups the overall training.

\begin{table}[thb]
\centering
\medskip
\caption{Classification mode comparison for RandomNEG. The first three row results indicated in italic are from Table 1. Bold text shows the proposed RandomNEG model balancing accuracy and speed.}
\begin{tabular}{|l|l|r|r|r|}
\hline
\textbf{Model}          & \textbf{Implementation}   & \textbf{Training Time (s)}    & \textbf{Test Accuracy (\%)}   \\ \hline
                        & Sequential                & 7,178.71                      & 98.33                         \\
\textit{RandomNEG-Goodness }   & Single-Layer              & 1,974.15                      & 98.26                         \\       
                        & All-Layers                & 2,008.25                      & 98.17                         \\ \hline
                        & Sequential                & 8,104.96                      & 98.48                         \\
{RandomNEG-Softmax }    & Single-Layer              & 1,891.86                      & 98.31                         \\       
                        & {All-Layers}       & \textbf{1,786.30}             & \textbf{98.33}                \\ \hline
\end{tabular}
\label{tab:smax-random}
\end{table}


\subsection{Evaluation of the Performance-Optimized Model}
\label{newgood}
This section explores the evaluation of our Performance-Optimized Model that is created by incorporating a new goodness function into the Forward-Forward algorithm. Our analysis shows that while the Performance-Optimized Model performs slightly worse than the best-performing AdaptiveNEG Goodness model, there is a notable reduction in training times, indicating a significant step towards efficient computing without significantly sacrificing performance.

In the MNIST dataset, the results presented in Table \ref{tab:PON} highlight the effectiveness of the proposed new goodness function when integrated with the Forward-Forward algorithm. The highest test accuracy is achieved by the AdaptiveNEG Goodness model, reaching 98.52\%, which is slightly superior to the 98.48\% accuracy of the RandomNEG Softmax model. It's noteworthy that the Performance-Optimized Network, whether using only the last layer or all layers, shows a small decrease in accuracy. Specifically, using all layers improves accuracy slightly to 98.38\% compared to 98.30\% when only the last layer is used. These results underscore the robustness of the proposed goodness function, particularly when all layers are optimized, despite a considerable reduction in training time (4219.97 seconds) compared to the AdaptiveNEG Goodness model (11190.72 seconds).

\begin{table}[thb]
\centering
\label{tab:PON}
\caption {MNIST evaluation of Performance-Optimized Model with the baseline models from Table 1.}
\begin{tabular}{|l|r|r|}
\hline
\textbf{Model}                                   & \textbf{Training Time (s)}    & \textbf{Test Accuracy (\%)}  \\ \hline
\textit{AdaptiveNEG-Goodness}                             & 11,190.72                     & 98.52                        \\ \hline
\textit{RandomNEG-Softmax}                                & 8,104.96                      & 98.48                        \\ \hline
Performance-Optimized Network (only last layer)  & 4,219.97                      & 98.30                        \\ \hline
Performance-Optimized Network (using all layers) & 4,219.97                      & 98.38                        \\ \hline
\end{tabular}
\end{table}

\newpage
\subsection{Experiments with CIFAR-10}
\label{cıfar}
To extend our analysis to a more complex scenario, we replicated the experiments described in sections \ref{main} through \ref{newgood} using the CIFAR-10 dataset. 
CIFAR-10, known for its higher variability and complexity compared to MNIST, serves as a challenging benchmark for testing the robustness and effectiveness of our models. 

Table \ref{tab:cifar} presents the detailed results of these experiments. The Performance-Optimized Network, whether using only the last layer or utilizing all layers, demonstrated a superior ability to handle the complexities of CIFAR-10, with the all-layers implementation achieving the highest accuracy of 53.50\%. This was closely followed by the RandomNEG Softmax model, which achieved 52.18\% accuracy. 
Furthermore, the Performance-Optimized Network models shows relatively efficient training durations compared to 
other PFF Models. 
Surprisingly, the AdaptiveNEG Goodness model, which performed exceptionally well on MNIST, significantly underperformed on CIFAR-10, with an accuracy of only 11.10\%. This difference suggests a potential mismatch between the model's capabilities and the dataset's requirements. 

It should be noted that while state-of-the-art models have achieved beyond 99\% accuracy on CIFAR-10, the performance of our models are similar to Hinton's original FF Algorithm, which reported an accuracy of 56\% using CNNs \cite{hinton2022forward}. 

\begin{table}[thb]
\centering
\caption {CIFAR-10 Results}
\label{tab:cifar}
\begin{tabular}{|l|r|r|}
\hline
\textbf{Model}                                      & \textbf{Training Time (s)}    & \textbf{Test Accuracy (\%)}   \\ \hline
Performance-Optimized Network (using all layers)    & \textbf{4,920.97}                      & \textbf{53.50}                \\ \hline
Performance-Optimized Network (only last layer)     & 4,920.97                      & 53.11                         \\ \hline
FixedNEG-Softmax                                   & 8,021.15                      & 50.89                         \\ \hline
RandomNEG-Softmax                                   & 7,636.99                      & 52.18                         \\ \hline
AdaptiveNEG-Goodness                                & 10,148.23                     & 11.10                         \\ \hline

\end{tabular}
\end{table}

\section{Conclusion and Future Work}
Our work in this paper presents Pipeline Forward-Forward Algorithm (PFF), a novel way of training distributed neural networks using Forward-Forward Algorithm. Compared to the classic implementations with backpropagation and pipeline parallelism \cite{huang2019gpipe} \cite{pipedream}, PFF is inherently different as it does not enforce the dependencies of backpropagation to the system, thus achieving higher utilization of computational units with less bubbles and idle time. Experiments that are done with PFF shows that PFF Algorithm achieves the same accuracy as standard FF implementation \cite{hinton2022forward} with a 4x speed up. Comparison of PFF with an existing distributed implementation of Forward-Forward (DFF \cite{deng2023dff}) shows even greater improvements as PFF achieves \%5 more accuracy in 10 times less epochs. This improvement in accuracy is mainly because PFF splits the data into batches and feeds it to the network batch by batch unlike DFF which feeds the data as whole. In addition, the data that is exchanged between layers in PFF is a lot less than DFF since PFF sends the layer information (weights and biases) whereas DFF sends the whole output of the data. This results in less communication overhead compared to DFF. 

Besides the exciting results that PFF produced, we believe that our work paves the way for a brand new path in the area of Distributed training of Neural Networks. Thus, there are plenty of different ways that can improve PFF, with some of these approaches outlined below.

\begin{itemize}
    \item \textbf{Exchanging parameters after each batch:} In the current implementation of PFF, exchange of parameters between different layers is done after each chapter. Making this exchange after each batch is worth trying since it could tune the weights better and produce higher accuracy. However, it could potentially increase the communication overhead. 
    \item \textbf{Usage of PFF in Federated Learning:} As PFF trains the model by only exchanging layer information between different nodes, it can be implemented to create a Federated Learning system where each node participates with their own data without sharing it with the other nodes.
    \item \textbf{Different Setups:} In the experiments of this work, we used sockets to establish communication between different nodes. This brings additional communication overhead since the data is sent through network. In a setup where computational units of the PFF are closer and can access to a shared resource (Multi GPU architectures) the time of training networks can decrease drastically.
    \item \textbf{Generating Negative Samples Differently:} The way that negative samples are generated is a very important aspect of Forward-Forward Algorithm as it directly affects the learning of the network. Thus, discovering new and better ways of generating negative samples would definitely result in a better performing system.
    \item \textbf{Forming an Innovative Framework:} A general and efficient framework for training large neural networks can be realized following and improving the novel ideas that are presented in this paper.
\end{itemize}

\newpage
\bibliographystyle{unsrt}  
\bibliography{references}  

\end{document}